\documentclass[a4]{IEEEtran}

\usepackage{xeCJK}
\newcommand{\CJK}[1]{{\CJKfamily{min}#1}}

\usepackage{color}

\usepackage{enumitem}

\usepackage{amsmath}

\usepackage{subcaption}

\usepackage{algorithm}
\usepackage{algpseudocode}
\definecolor{comment}{rgb}{0,0.6,0.3}

\usepackage{xurl}
\usepackage[hidelinks, colorlinks=true, urlcolor=blue, citecolor=comment]{hyperref}

\usepackage[capitalise]{cleveref}

\usepackage{graphicx}
\graphicspath{{./fig/}}

\begin{document}

\title{JSynFlow:\\Japanese Synthesised Flowchart Visual Question Answering Dataset\\built with Large Language Models}
\author{Hiroshi Sasaki / \href{https://www.jri.co.jp/en/}{The Japan Research Institute, Limited} /  \href{mailto:sasaki.hiroshi@jri.co.jp}{\url{sasaki.hiroshi@jri.co.jp}}
\thanks{
This article is a revised and translated version of a paper originally published in Japanese~\cite{sasaki2025jsynflow}.
}
\thanks{
Notice for the use of this material. The copyright of this material is retained by
the Japanese Society for Artificial Intelligence (JSAI). This material is published
here with the agreement of JSAI. Please be complied with Copyright Law of Japan
if any users wish to reproduce, make derivative work, distribute or make available
to the public any part or whole thereof.
All Rights Reserved, Copyright (C) The Japanese Society for Artificial Intelligence. 
}
}

\maketitle

\begin{abstract}
  Vision and language models (VLMs) are expected to analyse complex documents, such as those containing flowcharts, through a question-answering (QA) interface. The ability to recognise and interpret these flowcharts is in high demand, as they provide valuable insights unavailable in text-only explanations. However, developing VLMs with precise flowchart understanding requires large-scale datasets of flowchart images and corresponding text, the creation of which is highly time-consuming. To address this challenge, we introduce ``JSynFlow'', a synthesised visual QA dataset for Japanese flowcharts, generated using large language models (LLMs). Our dataset comprises task descriptions for various business occupations, the corresponding flowchart images rendered from domain-specific language (DSL) code, and related QA pairs. This paper details the dataset's synthesis procedure and demonstrates that fine-tuning with JSynFlow significantly improves VLM performance on flowchart-based QA tasks.
  Our dataset is publicly available at 
  \href{https://huggingface.co/datasets/jri-advtechlab/jsynflow}{\url{https://huggingface.co/datasets/jri-advtechlab/jsynflow}}.
\end{abstract}


\section{Introduction}

In recent years, vision-and-language models (VLMs), a class of large language models (LLMs) endowed with the capability to comprehend and interpret visual information, have witnessed rapid advancements. These models have enabled many advanced computer vision (CV) tasks such as visual content description through a simple question-and-answer (QA) style. Such CV tasks include the comprehension of flowchart images, which serve as visual representations of complex procedures and branching conditions. Flowcharts are extensively employed for business purposes, including process understanding, knowledge sharing, and workflow optimisation, making the ability of VLMs to recognise them in high demand. Nevertheless, substantial challenges persist in the development of VLMs capable of accurately interpreting the fine-grained visual details indispensable for the comprehension of flowcharts. Moreover, the acquisition of sufficiently large volumes of high-quality data, such as text-image aligned pairs for VLM training, remains a significant issue~\cite{li2025benchmark}.

\begin{figure*}[tbh]
   \begin{center}
     \begin{tabular}{cc}
      \hspace{-16mm}
       \begin{minipage}{0.7\hsize}
       \centering
       \begin{minipage}{0.5\hsize}
         \centering
         \includegraphics[clip,width=1.0\textwidth]{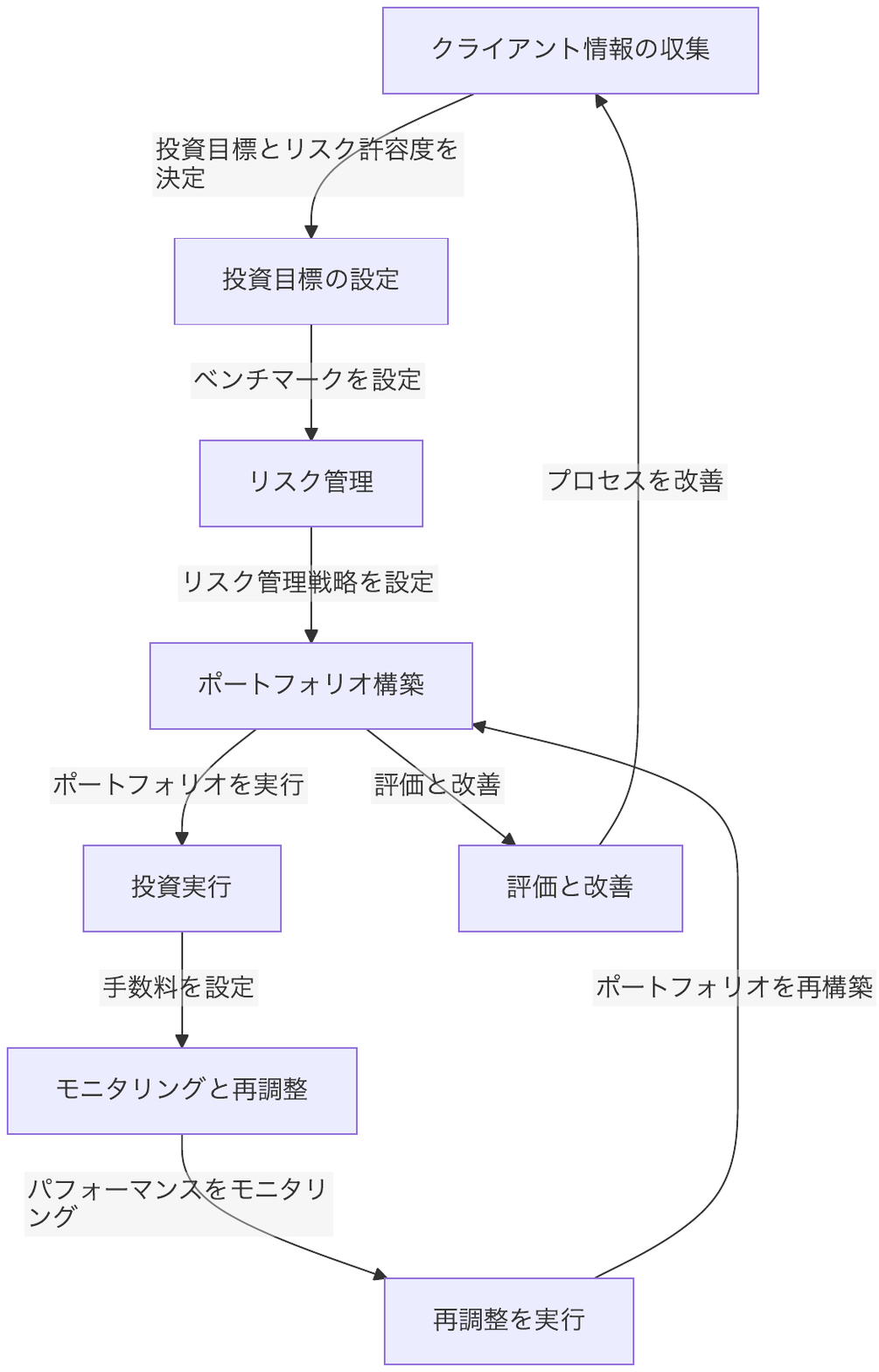}
      \end{minipage}
      \hspace{-7mm}
      \begin{minipage}{0.5\hsize}
         \includegraphics[clip,width=1.0\textwidth]{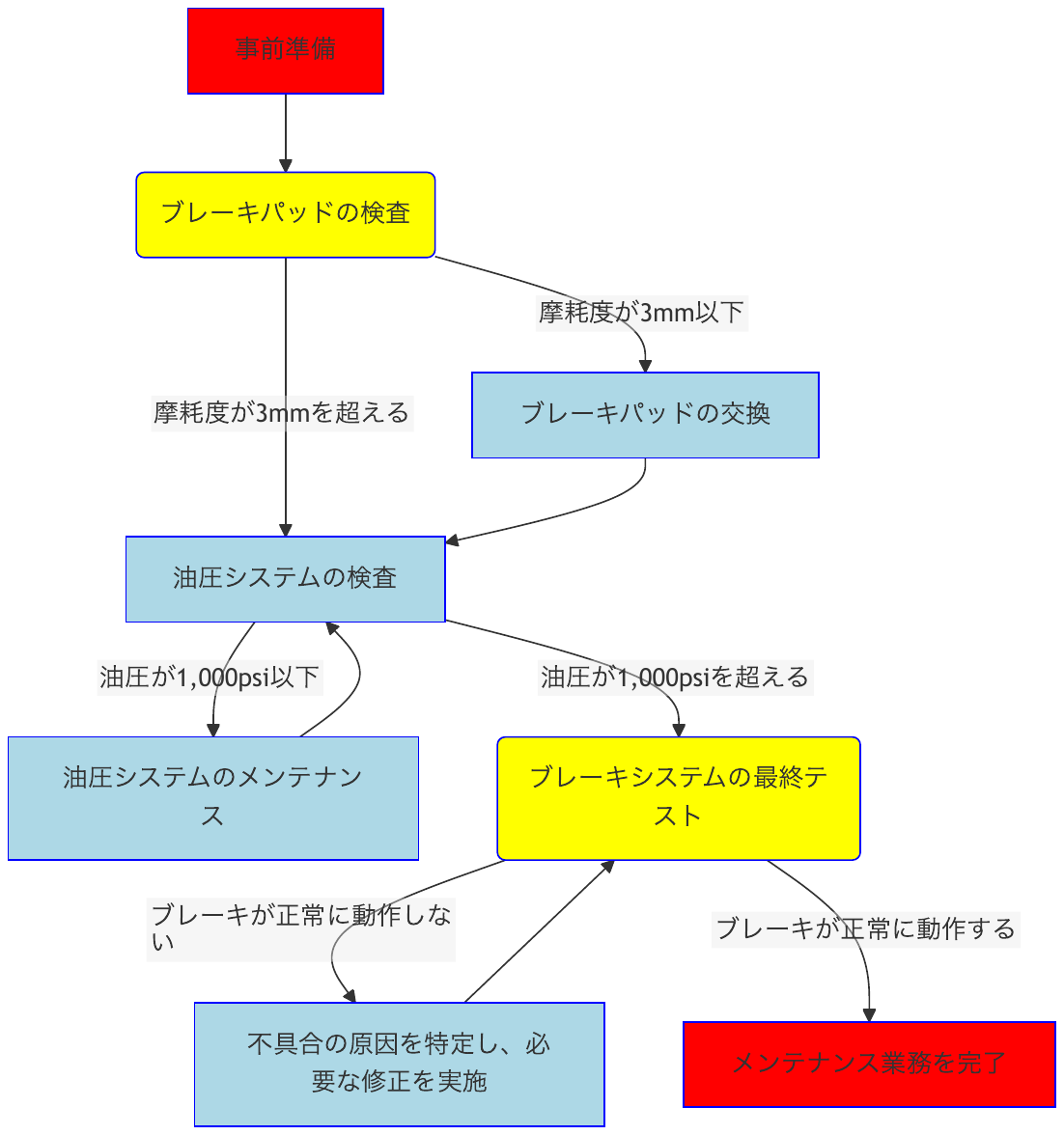}
      \end{minipage}
         \subcaption{Flowchart images}
         \label{fig:jsynflowimages}
       \end{minipage} 
          &
       \hspace{-5mm}
       \begin{minipage}{0.4\hsize}
         \vspace{-6mm} 
         \centering
           \includegraphics[clip,width=1.0\textwidth]{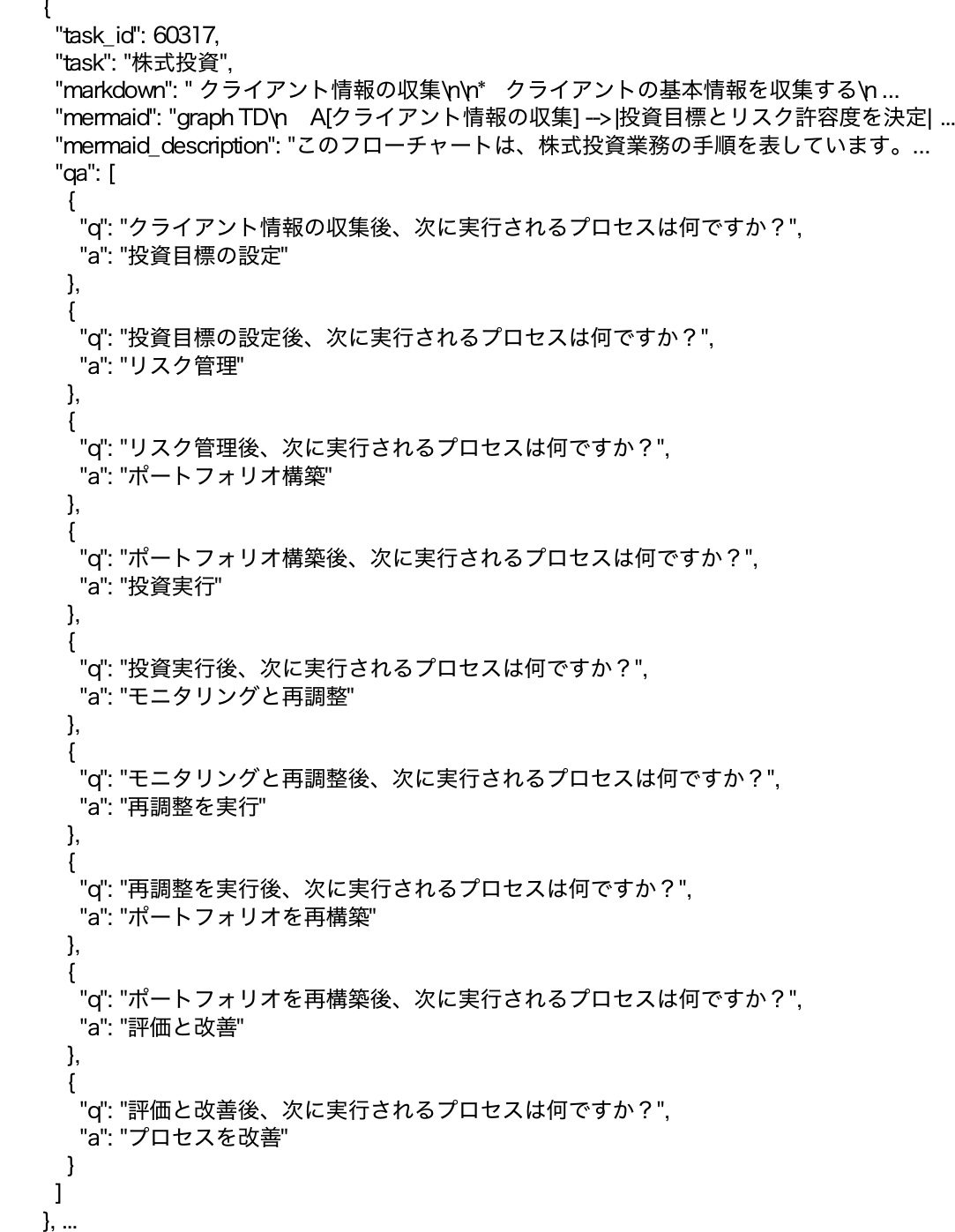}
           \subcaption{QA text}
           \label{fig:jsynflowjson}
         \end{minipage} \\
     \end{tabular}
     \caption{Our constructed Japanese flowchart VQA dataset}
     \label{fig:jflowvqa}
   \end{center}
 \end{figure*}

This paper introduces a Japanese flowchart dataset tailored to the visual question answering (VQA) task, constructed by leveraging LLMs (\cref{fig:jflowvqa}). We detail both the dataset itself and its construction methodology. Furthermore, to demonstrate its practical application, we present experimental results from fine-tuning VLMs with the dataset, illustrating its positive impact on QA performance.

\begin{table}
   \centering
   \caption{Statistics of the JSynFlow dataset}
   \label{tab:jflowvqa}
   \begin{tabular}{crrr} 
   \hline
         & Training & Test & Total \\ 
   \hline\hline
   Industries & &     & 9    \\
   Occupations  & &    & 115   \\
   Tasks & 1,359  & 152 & 1,511  \\
   QAs & 10,072  & 1,065 & 11,137  \\
   \hline
   \end{tabular}
\end{table}

\section{Related Work}

Domain-specific languages (DSLs) enable LLMs to produce text-based diagram definitions and generate the corresponding diagram images.
Representative examples of such DSLs for describing diagrams are PlantUML~\cite{std:plantuml}, primarily designed for UML diagrams, and Mermaid~\cite{std:mermaid}, which, in addition to UML, also supports flowcharts and graphs. 

Prior work on diagram dataset construction via DSL code generation includes approaches such as generating Mermaid code from publicly available procedural documents using LLMs~\cite{singh2024flowvqa}, and constructing datasets that combine randomly generated words with the rule-based generation of Mermaid code~\cite{pan2024flowlearn}.

\section{JSynFlow Dataset}
\label{sec:jflowvqa}
This section introduces our newly constructed Japanese flowchart VQA dataset, {\it JSynFlow} (\cref{fig:jflowvqa}).

\subsection{Overview}
The dataset is designed based on various occupations across multiple industries and their corresponding task procedures. It contains flowchart images representing each task (\cref{fig:jsynflowimages}) along with related textual data, which include procedural descriptions, DSL code representing the flowcharts, and QA pairs, in JSON format (\cref{fig:jsynflowjson}). The statistics of the dataset are shown in \cref{tab:jflowvqa}.

\subsection{Generation Procedure}
The dataset construction procedure is as follows:
\begin{enumerate}[label=(\roman*)]
   \item {\bf Generation of task lists (\cref{alg:tasklist})} \
   First, we use an LLM to generate a list of industry names and their corresponding occupations.
   Examples of the generated occupations are shown in \cref{tab:jobs}. Next, we use the LLM to generate a task list for each occupation.
   
   \item {\bf Generation of task procedures (\cref{alg:task})} \
   For each task, we generate a detailed procedural description using the LLM. Subsequently, we prompt the LLM to identify any ambiguities in the generated descriptions and revise them. This iterative process yields more refined and detailed procedural descriptions.
   
   \item {\bf Generation of flowcharts (\cref{alg:flowchart})} \
   From each procedural description, we use the LLM to generate DSL code that defines a flowchart. Since the generated code occasionally contains syntactic errors, any code that fails the image rendering of the DSL code (hereafter referred to as ``compilation'') is corrected by prompting the LLM with both the erroneous code and examples of correct syntax.
   
   \item {\bf Generation of QA pairs (\cref{alg:qa})} \
   Finally, for each successfully compiled DSL code, we use the LLM to generate corresponding question-and-answer (QA) pairs.
\end{enumerate}

\begin{table}
   \centering
   \caption{Examples of LLM-generated industries and occupations.}
   \label{tab:jobs}
   \small
   \begin{tabular}{cl} 
   \hline
   \hspace{-4mm}Industries      & \multicolumn{1}{c}{Occupations} \\ 
   \hline\hline
   \hspace{-3mm}\begin{tabular}{c}金融\\(Finance)\end{tabular} & \hspace{-8mm}\begin{tabular}{c}\CJK{インベストメントバンカー} (Investment banker),\\フィナンシャルプランナー (Financial Planner)\end{tabular}\\ \hline
   \hspace{-3mm}IT & \hspace{-5mm}\begin{tabular}{c}プログラマ(Programmer),\\ネットワークエンジニア (Network enginner)\end{tabular} \\ \hline
   \hspace{-3mm}\begin{tabular}{c}医療\\(Medical)\end{tabular} & \begin{tabular}{c}医師 (Doctor), 薬剤師 (Chemist),\\看護師 (Nurse),\\理学療法士 (Physiotherapist),\\栄養士 (Nutritionist), 歯科医師 (Dentist)\end{tabular} \\ \hline
   \hspace{-3mm}\begin{tabular}{c}法律\\(Legal)\end{tabular} & \begin{tabular}{c}裁判官 (Judge), 検察官 (Prosecutor),\\弁護士 (Barrister),\\司法書士 (Judicial Scrivener),\\弁理士 (Patent Attorney),\\裁判所職員 (Court Staff)\end{tabular} \\
   \hline
   \end{tabular}
\end{table}

\begin{algorithm}[tbh]
	\caption{Generation of task lists}
	\label{alg:tasklist}
	\renewcommand{\algorithmicrequire}{\textbf{Input}}
	\renewcommand{\algorithmicensure}{\textbf{Output}}
   \newcommand{\Desc}[2]{\Statex \makebox[6em][l]{#1:}#2}
   \algrenewcomment[1]{// #1}
   \algnewcommand{\LineComment}[1]{\State \color{comment} ~//#1\color{black}}
   \algnewcommand{\LineDesc}[2]{\makebox[5em][l]{#1}:~#2}
   \algnewcommand{\Initialize}[1]{\State {\bf Initialization}:~#1}
   \small
	\begin{algorithmic}[1]
		\Require $p^{\textnormal{*}}$: LLM prompt（see \cref{tab:prompt}）
		\Ensure $N^t $: Number of total tasks,
      \Statex $T=\{\left(j_i, t_i\right)\}^{N^t}_{i=1}$: List of all tasks across occupations
      \LineComment{1. Generate Industries \& Representative Occupations}
      \LineComment{ $N^g$: Number of industries}
      \LineComment{ $\{g_i\}^{N^g}_{i=1}$: Industries}
      \LineComment{ $\{j_{i,1}\}^{N^g}_{i=1}$: Representative occupations}
      \LineComment{ $\textnormal{LLM}(\cdot)$: LLM execution}
		\State $\left(N^g, \{g_i\}^{N^g}_{i=1}, \{j_{i,1}\}^{N^g}_{i=1}\right) \gets \textnormal{LLM}(p^\textnormal{groups})$;
		\LineComment{2. Generate Occupations per Industry}
      \LineComment{ $N^j_i$: Number of occupations for industry $g_i$}
      \LineComment{ $\{j_{i,k}\}^{N^j_i}_{k=1}$: List of occupations for industry $g_i$}
      \For{$i \gets 1$ to $N^g$}
         \State $\left(N^j_i, \{j_{i,k}\}^{N^j_i}_{k=2}\right) \gets \textnormal{LLM}(p^\textnormal{jobs}(g_i, j_{i,1}))$;
      \EndFor
      \LineComment{3. Generate Tasks per Occupation}
      \LineComment{ $N^t_{i,k}$: Number of tasks for occupation $j_{i,k}$}
      \LineComment{ $\{t_{i,k,l}\}^{N^t_{i,k}}_{l=1}$: List of tasks for occupation $j_{i,k}$}
      \Initialize{$N^t \gets 0$, $T \gets \Phi$}
      \For{$i \gets 1$ to $N^g$}
         \For{$k \gets 1$ to $N^j_i$}
            \State $\left(N^t_{i,k}, \{t_{i,k,l}\}^{N^t_{i,k}}_{l=1}\right) \gets \textnormal{LLM}(p^\textnormal{tasks}(j_{i,k}))$;
            \State $N^t \gets N^t + N^t_{i,k}$, $T \gets T \cup \{\left(j_{i,k}, t_{i,k,l}\right)\}^{N^t_{i,k}}_{l=1}$;
         \EndFor
      \EndFor
	\end{algorithmic}
\end{algorithm}

\begin{algorithm}[thb]
	\caption{Generation of task procedures}
	\label{alg:task}
	\renewcommand{\algorithmicrequire}{\textbf{Input}}
	\renewcommand{\algorithmicensure}{\textbf{Output}}
   \newcommand{\Desc}[2]{\Statex \makebox[5em][l]{#1:}#2}
   \algrenewcomment[1]{\(\triangleright\) #1}
   \algnewcommand{\LineComment}[1]{\State \color{comment} ~//#1\color{black}}
   \algnewcommand{\LineDesc}[2]{\makebox[3em][l]{#1}:~#2}
   \small
	\begin{algorithmic}[1]
		\Require $p^{\textnormal{*}}$: LLM prompt (see \cref{tab:prompt}),
         \Statex $\{\left(j_i, t_i\right)\}^{N^t}_{i=1}$: List of tasks,
         $N^r$: Number of revision iterations
		\Ensure $\{d^{N^r}_i\}^{N^t}_{i=1}$: Final task procedures
      \LineComment{1. Initial generation of task procedures}
      \For{$i \gets 1$ to $N^t$}
         \State $d^0_i \gets \textnormal{LLM}(p^\textnormal{desc}(j_i, t_i))$;
      \EndFor
      \LineComment{2. Task procedure revision}
      \LineComment{ $r^j_i$: Identified ambiguities in $d^j_i$}
      \For{$k \gets 0$ to $N^r-1$}
         \For{$i \gets 1$ to $N^t$}
            \State $r^k_i \gets \textnormal{LLM}(p^\textnormal{review}(j_i, t_i, d^k_i))$;
            \State $d^{k+1}_i \gets \textnormal{LLM}(p^\textnormal{revise}(j_i, t_i, d^k_i, r^k_i))$;
         \EndFor
      \EndFor
	\end{algorithmic}
\end{algorithm}

\begin{algorithm}[tbh]
	\caption{Generation of flowcharts}
	\label{alg:flowchart}
	\renewcommand{\algorithmicrequire}{\textbf{Input}}
	\renewcommand{\algorithmicensure}{\textbf{Output}}
   \newcommand{\Desc}[2]{\Statex \makebox[5em][l]{#1:}#2}
   \algrenewcomment[1]{\(\triangleright\) #1}
   \algnewcommand{\LineComment}[1]{\State \color{comment} ~//#1\color{black}}
   \algnewcommand{\LineDesc}[2]{\makebox[4em][l]{#1}:~#2}
   \algnewcommand{\Initialize}[1]{\State {\bf Initialization}:~#1}
   \small
	\begin{algorithmic}[1]
		\Require $p^{\textnormal{*}}$: LLM prompt (see \cref{tab:prompt}),
         \Statex $\{\left(j_i, t_i\right)\}^{N^t}_{i=1}$: List of all tasks across occupations,
         \Statex $\{d^{N^r}_i\}^{N^t}_{i=1}$: Task procedures,
         \Statex $N^c$: Number of DSL code revision iterations
		\Ensure $\{c^{N^c}_i\}^{N^t}_{i=1}$: Final DSL codes or $\phi$ (generation failure),
         \Statex $\{I_i\}^{N^t}_{i=1}$: Final flowchart images or $\textnormal{ERR}$ (compilation failure)
      \LineComment{1. Initial DSL Code Generation}
      \For{$i \gets 1$ to $N^t$}
		      \State $c^0_i \gets \textnormal{LLM}(p^\textnormal{DSL}(j_i, t_i, d^{N^r}_i))$;
      \EndFor
		\LineComment{2. Iterative DSL Code Compilation and Revision}
      \LineComment{ $\textnormal{DSL}(\cdot)$: DSL compliler function}
      \Initialize{$C^v \gets \Phi, C^e \gets \Phi$}
      \For{$k \gets 0$ to $N^c-1$}
         \For{$i \gets 1$ to $N^t$}
            \State $I_i \gets \textnormal{DSL}(c^k_i)$;
            \If{$I_i = \textnormal{ERR}$}
               \State $C^e \gets C^e \cup c^l_i$;
               \If{$k < N^c-1$}
                  \State $c^{k+1}_i \gets \textnormal{LLM}(p^\textnormal{DSLrev}(c^k_i, c\sim C^v))$;
               \Else
                  \State $c^{k+1}_i \gets \phi$;
               \EndIf
            \Else
               \State $C^v \gets C^v \cup c^k_i$;
               \State $c^{k+1}_i \gets c^k_i$;
            \EndIf
         \EndFor
      \EndFor
	\end{algorithmic}
\end{algorithm}

\begin{algorithm}[tbh]
	\caption{Generation of QA pairs}
	\label{alg:qa}
	\renewcommand{\algorithmicrequire}{\textbf{Input}}
	\renewcommand{\algorithmicensure}{\textbf{Output}}
   \newcommand{\Desc}[2]{\Statex \makebox[7em][l]{#1:}#2}
   \algrenewcomment[1]{\(\triangleright\) #1}
   \algnewcommand{\LineComment}[1]{\State \color{blue} ~//#1\color{black}}
   \algnewcommand{\LineDesc}[2]{\makebox[4em][l]{#1}:~#2}
   \small
	\begin{algorithmic}[1]
		\Require $p^{\textnormal{*}}$: LLM prompt (see \cref{tab:prompt}),
         \Statex $\{\left(j_i, t_i\right)\}^{N^t}_{i=1}$: Task list,
         \Statex $\{c^{N^c}_i\}^{N^t}_{i=1}$: DSL codes or invalid symbol ($\phi$)
		\Ensure $\{N^q_i\}^{N^t}_{i=1}$: Number of QA pairs for $t_i$ or $\phi$,
         \Statex $\{\{q_{i,k}\}^{N^q_i}_{k=1}\}^{N^t}_{i=1}$: QA pairs for $t_i$ or $\phi$
      \For{$i \gets 1$ to $N^t$}
         \If{$c^{N^c}_i \neq \phi$}
		      \State $\left(N^q_i, q_{i,k}\right) \gets \textnormal{LLM}\left(p^\textnormal{QA}(j_i, t_i, c^{N^c}_i)\right)$;
         \Else
            \State $\left(N^q_i, q_{i,k}\right) \gets (\phi,\phi)$
         \EndIf
      \EndFor
	\end{algorithmic}
\end{algorithm}

We used Llama-3.1-405B-Instruct~\cite{abhimanyu2024llama, llama3.1:405b} as the LLM ans Mermaid as the DSL. The mermaid-cli~\cite{code:mermaid-cli} tool was utilised to compile the DSL code into images. We set $N^r = N^c = 1$.

The generated flowcharts and QA pairs for each task were randomly divided into training and evaluation datasets, according to the sample sizes detailed in~\cref{tab:jflowvqa}.
Since some QA pairs were invalid due to LLM hallucinations, they were manually removed or corrected.

\subsection{Application to VLM Fine-Tuning}
To demonstrate a practical application of the dataset, we conducted fine-tuning experiments on several VLMs. We applied LoRA~\cite{hu2022lora} fine-tuning to two models: LLaVA-JP~\cite{llava:jp-1.3b-v1.1}, a variant of LLaVA~\cite{liu2024visual} previously fine-tuned on Japanese data, and Qwen2-VL-2B-Instruct~\cite{qwen:2-vl-2b-instruct}, which is based on Qwen2-VL~\cite{wang2024qwen2-vl}, using the training set of flowchart images and QA pairs from our dataset. The training followed the default parameter settings provided in the publicly available fine-tuning scripts\footnote{\url{https://github.com/tosiyuki/LLaVA-JP/blob/v1.1.0/scripts/finetune/finetune_lora_llm_jp.sh}}$^,$\footnote{\url{https://github.com/2U1/Qwen2-VL-Finetune/blob/62c14c08892e56a3aed0790d5d1be71b1f9dca8a/scripts/finetune_lora.sh}}, with the exception that the number of epochs was set to 3.

For evaluation, we used the test set from the dataset. For each QA pair, we input the flowchart image and its associated question into the model. We then calculated the similarity between the model's predicted answer and the ground-truth answer using BERTScore~\cite{zhang2020bertscore}. A multilingual BERT model~\cite{bert:base-multilingual-cased} was used for this calculation. Finally, we compared the average BERTScore before and after fine-tuning. The results, presented in~\cref{tab:bertscore}, demonstrate that fine-tuning on the proposed dataset yielded substantial performance improvements.

\begin{table}
   \centering
   \caption{Comparison of BERTScore~\cite{zhang2020bertscore} performance before (Baseline) and after fine-tuning with JSynFlow.}
   \label{tab:bertscore}
   \begin{tabular}{lccc}
   \hline
      & {\small Precision} & Recall & F1 \\
   \hline\hline
   \multicolumn{4}{l}{LLaVA} \\ 
   baseline & 0.6773 & 0.6546 & 0.6605 \\
    + JSynFlow  & {\bf 0.7627} & {\bf 0.7792} & {\bf 0.7691} \\
   \hline
   \multicolumn{4}{l}{Qwen2-VL} \\
   baseline & 0.8329 & 0.8578 & 0.8597 \\
    + JSynFlow  & {\bf 0.9372} & {\bf 0.9360} & {\bf 0.9397} \\
   \hline
   \end{tabular}
\end{table}

\section{Conclusion}
\label{sec:discussion}
In this study, we have detailed the construction of a Japanese flowchart VQA dataset using LLMs and demonstrated that fine-tuning VLMs with this dataset leads to significant performance improvements.

However, several limitations remain. The quality of the dataset is highly dependent on the LLM's output, which currently necessitates manual verification and correction. Furthermore, the visual appearance of the generated images, being outputs from a standard diagram renderer, exhibits less diversity than real-world data.

Future work will focus on addressing these limitations and establishing more robust methods for high-quality dataset synthesis.

\begin{table*}[tb]
   \centering
   \caption{Prompts used for LLM-based data generation. Refer to~\cref{sec:jflowvqa} for a detailed explanation of the variables in the first column and enclosed in $\langle \rangle$.}
   \label{tab:prompt}
   \begin{tabular}{cl} 
   \hline
      & \multicolumn{1}{c}{\underline{Actual prompt templates (in Japanese)}} \\
      & \multicolumn{1}{c}{English translation} \\ 
   \hline\hline
   $p^\textnormal{groups}$ &
   \hspace{-2mm}
   \small
   \begin{tabular}{l}
      高度で複雑な業務が求められる代表的な職業とその業界の例を、IT分野に偏らず、可能な限り多種多様に、多く\\の人が知る有名なもののみに限定し、各業界1件ずつ、合計10件挙げ、以下の例のようなCSV形式で結果のみ\\出力してください。\\\textasciigrave\textasciigrave\textasciigrave csv\\医師,医療業界\\プログラマ,IT業界\\\textasciigrave\textasciigrave\textasciigrave\\また、自信のない回答は出力しないでください。\\
      \hline
      Provide a list of 10 representative occupations that require advanced and complex tasks, along with their\\ respective industries, are as diverse as possible, not concentrated in the IT field, and limited to widely known\\ professions. The output should include only the result in CSV format like the following example. \\\textasciigrave\textasciigrave\textasciigrave csv\\Doctor,Medical Industry\\
      Programmer,IT Industry\\\textasciigrave\textasciigrave\textasciigrave\\If you are not confident in your answer, do not provide an output.
   \end{tabular}     \\
   \hline
   $p^\textnormal{jobs}$ &
   \hspace{-2mm}
   \small
   \begin{tabular}{l}
      $\langle j_{i,1}\rangle$は$\langle g_i\rangle$の代表的な職業ですが、$\langle g_i\rangle$の他の職業の例を、多くの人が知るような有名で、かつ高度で複雑な\\業務が求められるものに限定し、可能な限り多種多様に、2件以上かつ20件以下の範囲で列挙し、;で区切って\\結果のみ出力してください。回答に自信がない場合は出力をしないでください。\\
      \hline
      Given $\langle j_{i,1}\rangle$ is a representative occupation for $\langle g_i\rangle$, list other occupations within $\langle g_i\rangle$ that are limited to widely\\ known professions and require advanced and complex tasks. The output should include only the result of\\ a diverse list of at least 2, but no more than 20, occupations, separated by semicolons. If you are not confident\\ in your answer, do not provide an output.
   \end{tabular}     \\
   \hline
   $p^\textnormal{tasks}$ &
   \hspace{-2mm}
   \small
   \begin{tabular}{l}
      あなたは$\langle j_{i,k}\rangle$です。$\langle j_{i,k}\rangle$の高度で複雑な業務の例を、具体的に2件以上かつ20件以下の範囲で、可能な限り\\多種多様に列挙し、;区切りで結果のみ出力してください。あなたがこの職業について知らなかったり、具体的な\\業務の例を自信を持って挙げられない場合は何も出力しないでください。\\
      \hline
      You are a $\langle j_{i,k}\rangle$. Provide between 2 and 20 diverse examples of the advanced and complex tasks in the role of\\ a $\langle j_{i,k}\rangle$. The output should include only the result separated by a semicolon. If you are not confident in your\\ answer, do not provide an output.
   \end{tabular}     \\
   \hline
   $p^\textnormal{desc}$ &
   \hspace{-2mm}
   \small
   \begin{tabular}{l}
      あなたはベテラン$\langle j_i\rangle$です。新人に指導するため、「$\langle t_i\rangle$」業務のマニュアルを、条件分岐を含むような詳細かつ\\複雑な手順の一つ一つについて、各分岐条件を明確にして、箇条書きで簡潔に、Markdown形式で出力してくだ\\さい。あなたがこの業務について知らなかったり、自信を持って述べられない場合は何も出力しないでください。\\
      \hline
      You are an experienced $\langle j_i\rangle$. For a new employee, create a manual for the '$\langle t_i\rangle$' task describing each detailed\\ and complex step, including conditional branches, clearly specify each branching condition and output\\ the instructions concisely in a bulleted list using Markdown format. If you are not familiar with this task or\\ cannot describe it with confidence, do not provide an output.
   \end{tabular}    \\
   \hline
   $p^\textnormal{review}$ &
   \hspace{-2mm}
   \small
   \begin{tabular}{l}
      あなたは新人$\langle j_i\rangle$です。以下はベテラン$\langle j_i\rangle$により書かれた、「$\langle t_i\rangle$」業務のマニュアルで、その手順が\\Markdown記法で記述されています。\\\textasciigrave\textasciigrave\textasciigrave markdown\\$\langle d^k_i\rangle$\\\textasciigrave\textasciigrave\textasciigrave\\あなたはこのマニュアルに従い、「$\langle t_i\rangle$」業務を行うことを求められています。実際に業務を行うにあたり、この\\マニュアルに対する不明点や疑問点を挙げてください。あなたはMarkdown記法は熟知しているものとします。\\また、不明点や疑問点がない場合は何も出力しないでください。\\
      \hline
      You are a junior $\langle j_i\rangle$. Below is a manual for the '$\langle t_i\rangle$' task written by an experienced $\langle j_i\rangle$, in which the\\ procedure is described using Markdown format.
      \\\textasciigrave\textasciigrave\textasciigrave markdown\\$\langle d^k_i\rangle$\\\textasciigrave\textasciigrave\textasciigrave\\
      You are required to carry out the '$\langle t_i\rangle$' task in accordance with this manual. Before actually performing\\ the task, list any points that are unclear or any questions you have about this manual. You are fully familiar\\ with Markdown syntax. If you have no unclear points or questions, do not provide an output.
   \end{tabular}     \\
   \hline
   \multicolumn{2}{r}{Continued on next page} \\
   \end{tabular}
\end{table*}
\addtocounter{table}{-1}
\begin{table*}[tb]
   \caption{Prompts used for LLM-based data generation. Refer to~\cref{sec:jflowvqa} for a detailed explanation of the variables in the first column and enclosed in $\langle \rangle$. (Continued)}
   \begin{tabular}{cl} 
   \hline
   & \multicolumn{1}{c}{\underline{Actual prompt templates (in Japanese)}} \\
   & \multicolumn{1}{c}{English translation} \\ 
   \hline\hline
   $p^\textnormal{revise}$ &
   \hspace{-2mm}
   \small
   \begin{tabular}{l}
      あなたはベテラン$\langle j_i\rangle$です。以下は「$\langle t_i\rangle$」業務について、あなたが新人に指導するために書いたマニュアルで、\\その手順がMarkdown記法で記述されています。\\\textasciigrave\textasciigrave\textasciigrave markdown\\$\langle d^k_i\rangle$\\\textasciigrave\textasciigrave\textasciigrave\\このマニュアルについて、新人から以下のような質問がありました。\\---\\$\langle r^k_i\rangle$\\---\\これらの質問を解決するよう、マニュアルに記載されているMarkdown記法の手順を修正してください。\\あなたが自信を持って質問に対応できない場合は何も出力しないでください。\\
      \hline
      You are an experienced $\langle j_i\rangle$. The following is a manual you have written to guide a new enployee in\\ the $\langle t_i\rangle$ task, with the procedures described using Markdown format.\\
      \textasciigrave\textasciigrave\textasciigrave markdown\\
      $\langle d^k_i\rangle$\\
      \textasciigrave\textasciigrave\textasciigrave \\
      A new enployee has asked the following questions about this manual.\\
      ---\\
      $\langle r^k_i\rangle$\\
      ---\\
      Revise the Markdown procedures in the manual to address these questions. If you are not confident in\\ answering the questions, do not provide an output.
   \end{tabular}     \\
   \hline
   $p^\textnormal{DSL}$ &
   \hspace{-2mm}
   \small
   \begin{tabular}{l}
      あなたはベテラン$\langle j_i\rangle$です。新人に指導するため、「$\langle t_i\rangle$」業務の手順を、Markdown形式で以下の通り作成しま\\した。\\\textasciigrave\textasciigrave\textasciigrave markdown\\$\langle d^{N^r}_i\rangle$\\\textasciigrave\textasciigrave\textasciigrave \\これらの手順について、各手順と分岐条件を明確にして、フローチャートを作成してMermaid形式で出力して\\ください。あなたがこの手順についてよく理解できなかったり、自信を持って述べられない場合は何も出力\\しないでください。\\
      \hline
      You are an experienced $\langle j_i\rangle$. To guide a new employee, you have written the procedure for the '$\langle t_i\rangle$' task using\\ Markdown format, as shown below. \\
      \textasciigrave\textasciigrave\textasciigrave markdown\\$\langle d^{N^r}_i\rangle$ \\
      \textasciigrave\textasciigrave\textasciigrave \\
      Based on these instructions, generate a flowchart in Mermaid format, clearly defining each step and any\\ branching conditions. If you do not fully understand the procedure or are not confident in generating\\ the flowchart, do not provide an output.
   \end{tabular}     \\
   \hline
   $p^\textnormal{DSLrev}$ &
   \hspace{-2mm}
   \small
   \begin{tabular}{l}
      以下は正しいMermaid記法のフローチャートです。\\\textasciigrave\textasciigrave\textasciigrave mermaid\\$\langle c\rangle$\\\textasciigrave\textasciigrave\textasciigrave\\以下のMermaid記法のフローチャートは部分的に正しくない記法で記述されています。このMermaid記法の\\フローチャートを正しく修正して出力してください。\\\textasciigrave\textasciigrave\textasciigrave mermaid\\$\langle c^k_i\rangle$\\\textasciigrave\textasciigrave\textasciigrave \\
      \hline
      The following is a flowchart with correct Mermaid notation.\\
      \textasciigrave\textasciigrave\textasciigrave mermaid\\$\langle c\rangle$\\\textasciigrave\textasciigrave\textasciigrave\\
      The Mermaid flowchart below contains incorrect notation. Correct the notation and output the valid flowchart.\\
      \textasciigrave\textasciigrave\textasciigrave mermaid\\$\langle c^k_i\rangle$\\\textasciigrave\textasciigrave\textasciigrave
   \end{tabular}     \\
   \hline
   \multicolumn{2}{r}{Continued on next page} \\
   \end{tabular}
\end{table*}
\addtocounter{table}{-1}
\begin{table*}[tb]
   \caption{Prompts used for LLM-based data generation. Refer to~\cref{sec:jflowvqa} for a detailed explanation of the variables in the first column and enclosed in $\langle \rangle$. (Continued)}
   \begin{tabular}{cl} 
   \hline
   & \multicolumn{1}{c}{\underline{Actual prompt templates (in Japanese)}} \\
   & \multicolumn{1}{c}{English translation} \\ 
   \hline\hline
   $p^\textnormal{QA}$ &
   \hspace{-2mm}
   \small
   \begin{tabular}{l}
      あなたはベテラン$\langle j_i\rangle$です。以下はあなたが新人のために書いた、「$\langle t_i\rangle$」業務のMermaid記法のフローチャート\\です。\\\textasciigrave\textasciigrave\textasciigrave mermaid\\$\langle c^{N^c}_i\rangle$\\\textasciigrave\textasciigrave\textasciigrave\\新人の理解確認のためにこのフローチャートに関するクイズを５問以上作成してください。問題と回答のペアは\\以下の例のようなJSON形式で出力してください。\\\textasciigrave\textasciigrave\textasciigrave json\\\lbrack\\  {"q": "1+1は？", "a": "2"},\\  {"q": "Aの次のアルファベットは？", "a": "B"}\\\rbrack\\\textasciigrave\textasciigrave\textasciigrave\\問題は必ずフローチャートの内容から回答を導き出せるものにしてください。また、あなたと新人はMermaid記法\\を熟知しているものとし、Mermaid記法そのものを問う問題は作成しないでください。あなたが自信を持って問題\\を作成できない場合は何も出力しないでください。\\
      \hline
      You are an experienced $\langle j_i\rangle$. Below is the flowchart for the '$\langle t_i\rangle$' task you have written for a new employee using\\ Mermaid format.\\
      \textasciigrave\textasciigrave\textasciigrave mermaid\\$\langle c^{N^c}_i\rangle$\\\textasciigrave\textasciigrave\textasciigrave\\
      To test the new employee's understanding, create a quiz with at least 5 questions based on this flowchart.\\ The question and answer pairs should be output in JSON format, as in the following example:\\
      \textasciigrave\textasciigrave\textasciigrave json\\\lbrack\\  {"q": "What is 1+1?", "a": "2"},\\  {"q": "What is the letter after A?", "a": "B"}\\\rbrack\\\textasciigrave\textasciigrave\textasciigrave\\
      Each question must be answerable solely from the content of the flowchart. Furthermore, assume that both you\\ and the new employee are proficient with Mermaid syntax, and do not create questions about the syntax itself.\\ If you are not confident in creating these questions, do not provide an output.
   \end{tabular}     \\
   \hline
   \end{tabular}
\end{table*}

\bibliographystyle{ieee_fullname}
\bibliography{references}

\end{document}